\pdfoutput=1
\newcommand{\CLASSINPUTinnersidemargin}{0.75in}
\newcommand{\CLASSINPUToutersidemargin}{0.75in}
\newcommand{\CLASSINPUTtoptextmargin}{0.75in}
\newcommand{\CLASSINPUTbottomtextmargin}{0.75in}

\documentclass[10pt,conference,letterpaper,final]{IEEEtran}

\makeatletter
\def\mdseries@tt{m}
\def\mdseries@rm{m}
\makeatother

\renewcommand{\IEEEtitletopspaceextra}{0.25in}

\usepackage[T1]{fontenc}
\usepackage[utf8]{inputenc}

\usepackage[usenames,dvipsnames]{xcolor}
\usepackage{algorithmic}
\usepackage{amsmath}
\usepackage{amssymb}
\usepackage{array}
\usepackage[pdftex]{graphicx}
\usepackage{placeins}
\usepackage[obeyFinal]{todonotes}
\usepackage{url}
\usepackage{xspace}
\usepackage{libertine} %
\usepackage{tikz}
\usepackage{pgfmath}
\usepackage{pgffor}
\usetikzlibrary{bending, patterns, backgrounds, calc, positioning, arrows.meta, shapes, chains, shapes.multipart, shapes.geometric, arrows, decorations.markings, decorations.pathreplacing, decorations.text}
\usepackage{ifdraft}

\usepackage{xstring}

\usepackage{dblfloatfix}

\usepackage{subfig}
\usepackage{paralist}

\usepackage[style=numeric-comp,sorting=none,minbibnames=3]{biblatex}
\usepackage{xfrac}
\usepackage[detect-all,binary-units]{siunitx}
\usepackage[normalem]{ulem}

\usepackage[font=small]{caption}
\DeclareCaptionFont{white}{\color{white}}
\DeclareCaptionFormat{listing}{\colorbox{gray}{\parbox{\linewidth}{#1#2#3}}}
\usepackage[font=itshape]{quoting}

\usepackage{booktabs}

\usepackage{listings,lstautogobble}
\usepackage[newfloat,finalizecache=false,frozencache=true]{minted}
\setminted{fontsize=\footnotesize}
\setminted{breaklines=true}
\setminted{autogobble=true}
\setminted{frame=bottomline}
\setminted{fontfamily=courier}

\usepackage{hyperref}

\newcommand\myshade{85}
\colorlet{mylinkcolor}{violet}
\colorlet{mycitecolor}{YellowOrange}
\colorlet{myurlcolor}{Aquamarine}

\hypersetup{
  linkcolor  = mylinkcolor!\myshade!black,
  citecolor  = mycitecolor!\myshade!black,
  urlcolor   = myurlcolor!\myshade!black,
  colorlinks = true,
  breaklinks = true, %
}
\usepackage{breakurl} %

\usepackage{cleveref}
\crefname{lstlisting}{listing}{listings}
\Crefname{lstlisting}{Listing}{Listings}

\usepackage{microtype}

\makeatletter
\newcounter{IEEE@bibentries}
\renewcommand\IEEEtriggeratref[1]{%
  \renewbibmacro{finentry}{%
    \stepcounter{IEEE@bibentries}%
    \ifthenelse{\equal{\value{IEEE@bibentries}}{#1}}
    {\finentry\@IEEEtriggercmd}
    {\finentry}%
  }%
}
\makeatother

\newcommand{\codeInline}[1]{\textit{#1}}

\newcommand{\stackName}{BrainScaleS OS\xspace}
\newcommand{\stackNameExpand}{BrainScaleS Operating System}

\newcommand{\BSSWM}{BrainScaleS Wafer Module}

\newcommand{\BSS}[1]{BSS\nobreakdash-#1}
\newcommand{\BrainScaleS}[1]{BrainScaleS\nobreakdash-#1}

\newcommand{\drawsector}[6][]{%
        \def\rin{#2}
        \def\width{#3}
        \def\textr{\rin+.5*\width}
        \def\sangle{#4}
        \def\dangle{#5}
        \draw[#1] (\sangle:\rin) arc [start angle = \sangle, delta angle=-\dangle, radius=\rin]--++({\sangle-\dangle}:\width) arc [start angle = {\sangle-\dangle}, delta angle=\dangle, radius={\rin+\width}] --cycle;
        \pgfmathtruncatemacro\tsangle{\sangle-\dangle}
        \pgfmathtruncatemacro\teangle{\sangle}
        \ifnum \teangle>0
                \ifnum \tsangle>0
                        \pgfmathsetmacro\tsangle{\sangle}
                        \pgfmathsetmacro\teangle{\sangle-\dangle}
                \else \tsangle=0
                        \pgfmathsetmacro\tsangle{\sangle}
                        \pgfmathsetmacro\teangle{\sangle-\dangle}
                \fi
        \fi
        \draw[decorate,decoration={raise=-3pt, text along path, text=#6, text align={align=center}}] (\tsangle:\textr) arc(\tsangle:\teangle:\textr);
}

\begin{document}

\title{The Operating System of the Neuromorphic BrainScaleS-1 System}

\newcommand{\uheiSymbol}{\IEEEauthorrefmark{6}\xspace}
\newcommand{\tudSymbol}{\IEEEauthorrefmark{4}\xspace}

\author{
	\IEEEauthorblockN{%
		Eric Müller\IEEEauthorrefmark{1}\uheiSymbol,
		Sebastian Schmitt\IEEEauthorrefmark{1}\uheiSymbol,
		Christian Mauch\IEEEauthorrefmark{1}\uheiSymbol,\\
		Sebastian Billaudelle\uheiSymbol, %
		Andreas Grübl\uheiSymbol,      %
		Maurice Güttler\uheiSymbol,    %
		Dan Husmann\uheiSymbol,        %
		Joscha Ilmberger\uheiSymbol,   %
		Sebastian Jeltsch\uheiSymbol,\\
		Jakob Kaiser\uheiSymbol,       %
		Johann Klähn\uheiSymbol,       %
		Mitja Kleider\uheiSymbol,      %
		Christoph Koke\uheiSymbol,     %
		José Montes\uheiSymbol,        %
		Paul Müller\uheiSymbol,\\      %
		Johannes Partzsch\tudSymbol,  %
		Felix Passenberg\uheiSymbol,   %
		Hartmut Schmidt\uheiSymbol,    %
		Bernhard Vogginger\tudSymbol, %
		Jonas Weidner\uheiSymbol,\\    %
		Christian Mayr\tudSymbol
		and Johannes Schemmel\uheiSymbol}\\
	\IEEEauthorblockA{%
		\IEEEauthorrefmark{1}%
		contributed equally\\
		\uheiSymbol%
		Kirchhoff-Institute for Physics\\
		Ruprecht-Karls-Universität Heidelberg, Germany\\
		Email: \{%
			mueller,sschmitt,cmauch%
			\}@kip.uni-heidelberg.de
	}%
	\IEEEauthorblockA{%
		\tudSymbol%
		Chair of Highly-Parallel VLSI-Systems and Neuro-Microelectronics\\
		Technische Universität Dresden, Germany%
	}
}

\maketitle

\begin{abstract}
	\BrainScaleS{1} is a wafer-scale mixed-signal accelerated neuromorphic system targeted for research in the fields of computational neuroscience and beyond-von-Neumann computing.
The \stackNameExpand{} (\stackName{}) is a software stack giving users the possibility to emulate networks described in the high-level network description language PyNN with minimal knowledge of the system.
At the same time, expert usage is facilitated by allowing to hook into the system at any depth of the stack.
We present operation and development methodologies implemented for the \BrainScaleS{1} neuromorphic architecture and walk through the individual components of \stackName{} constituting the software stack for \BrainScaleS{1} platform operation.

\end{abstract}

\ifoptionfinal{} {\tableofcontents}

\section{Introduction}

State-of-the-art neuromorphic architectures pose many requirements in terms of system control, data preprocessing, data exchange and data analysis.
In all these areas, software is involved in satisfying these requirements.
Several neuromorphic systems are directly used by individual researchers in collaborations, e.g., \cite{furber2012overview,pfeil2013six,davies2018loihi,merolla2014million}.
In addition, some systems are operated as experiment platforms providing access for external users~\cite{furber2012overview,pfeil2013six,schemmel2010iscas,davies2018loihi}.

Especially systems open for a broader range of users require clear and concise interfaces.
Neuromorphic platform operators have additional requirements in resource management, runtime control and ---depending on data volumes--- ``grid-computing''-like data processing capabilities.
At the same time, usability and experiment reproducibility are crucial properties of all experiment platforms, including neuromorphic systems.

Modern software engineering techniques such as code review, continuous integration as well as continuous deployment can help to increase platform robustness and ensure experiment reproducibility.
Long-term hardware development roadmaps and experiment collaborations draw attention to platform sustainability.
Technical decisions need to be evaluated for potential future impact;
containing and reducing technical debt is a key objective during planning as well as development.
Regardless of being software-driven simulations\slash{}emulations, or being physical experiments, modern experiment setups more and more depend on these additional tools and skills in order to enable reproducible, correct and successful scientific research.

This paper describes the results of a ten-year project delivering the software environment and platform operation tools for the \BrainScaleS{1} neuromorphic system.
The following sections describe the hardware substrate and give a general overview.
\Cref{sec:meths} introduces the methods and software tools we employ.
In \cref{sec:impl}, the scopes and implementation details of the main software layers and libraries are explained, followed by an overview over the operation of the platform in \cref{sec:operation}.
\Cref{sec:app} exemplifies the usage of the \stackNameExpand{} on a simple experiment and describes larger experiments carried out in the past.
We close in \cref{sec:future} with an overview over future developments and discuss our endeavor and the lessons learned in \cref{sec:discussion}.

\subsection{The \protect\BrainScaleS{1} Neuromorphic System}\label{sec:bss1hw}

Classical neuromorphic systems make use of VLSI to implement electronic analog circuits mimicking neuro-biological architectures in the nervous system~\cite{mead90neuromorphic}.
Contemporary systems also employ mixed-signal techniques to enable flexible system connectivity based on conventional digital interfaces~\cite{moradi2014eventbased}.
Recently, purely digital systems emerged~\cite{furber2012overview,merolla2014million,davies2018loihi}.
Compared to the analog approach typical advantages of such systems are:
deterministic behavior and arbitrarily programmable neuron dynamics.
However, when comparing at the same technology node the energy as well as area consumption is increased.

Based on the ideas of a single-chip implementation called \textit{Spikey}~\cite{schemmel_ijcnn06}, \BSS{1} is a mixed-signal architecture providing accelerated Adaptive Exponential Integrate-and-Fire (AdEx) neuron dynamics and plastic synapses~\cite{millner2010vlsi,schemmel2010iscas,schmitt2016classification}.
While many neuromorphic systems target biological real-time execution (i.e.\ model time constants in the same order as their biological counterpart)~\cite{thakur2018mimicthebrain},
BSS-1 evolves in continuous time, typically at a speed-up factor of \num{1000}--\num{10000} faster than biological real time.
Consequently, real-time interfacing to, e.g., sensors and robotic applications are not the main goal of the architecture.
The design focuses on fast model dynamics, controllable model parameters and system scalability, thereby allowing for time-compressed emulations of longer experiment time scales.
Plasticity and learning processes can therefore be investigated in manageable time frames.

\begin{figure}[tbp]
  \hfill
  \subfloat[\label{wafer_module:exploded_view}]{\def\svgwidth{.5\linewidth}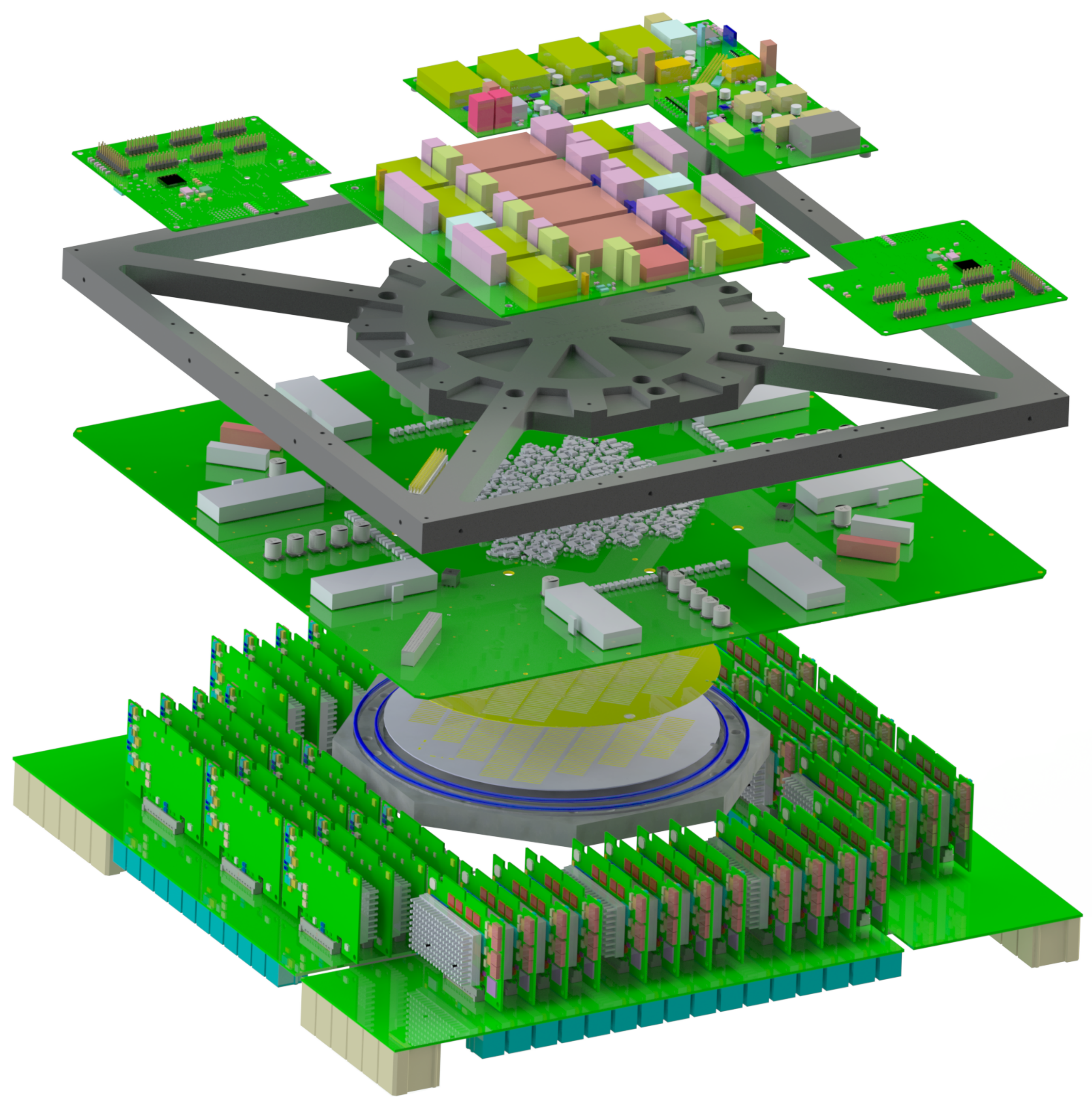}
  \hfill
  \subfloat[\label{wafer_module:picture}]{\includegraphics[height=0.2\textheight]{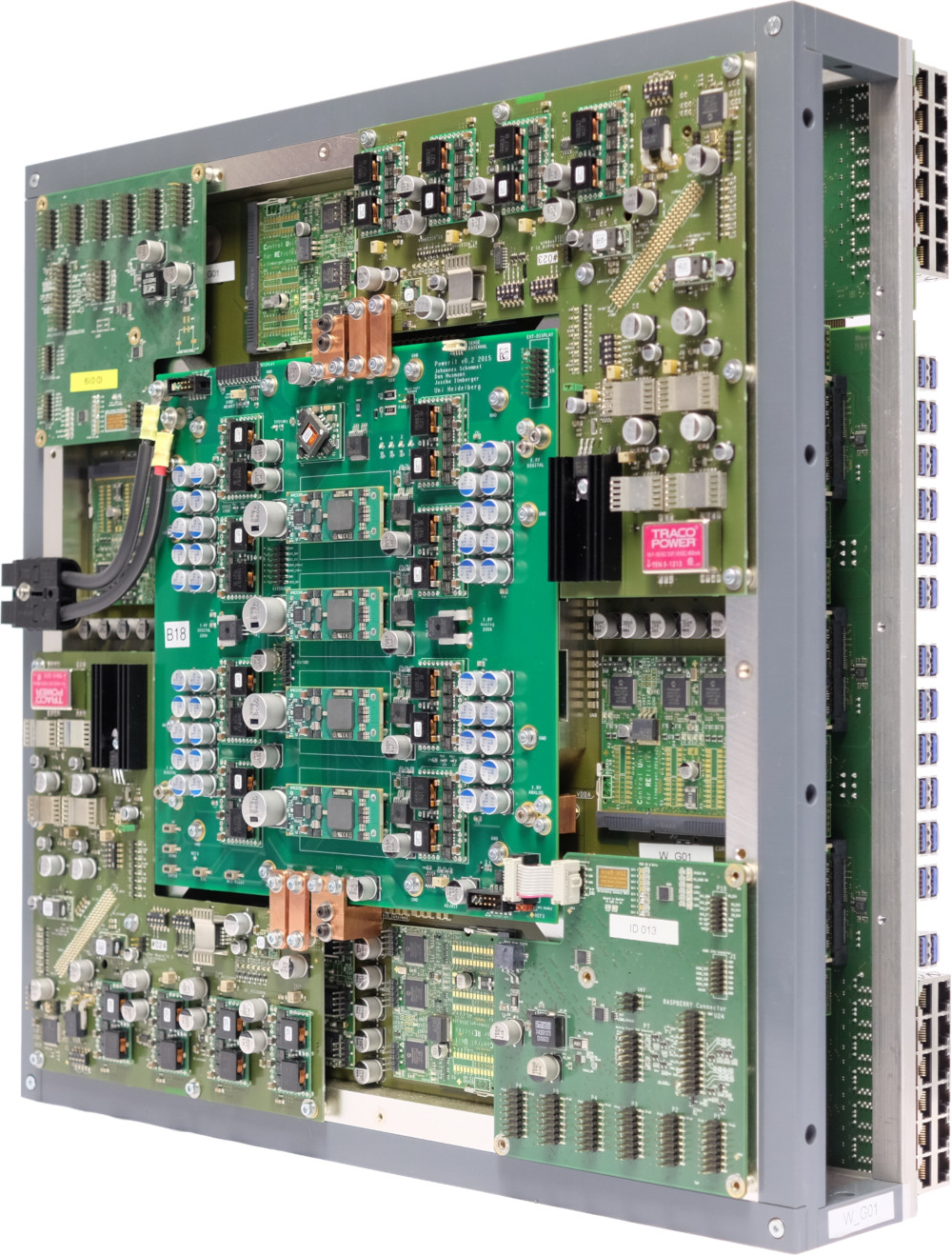}}
  \hspace*{\fill}
  \caption{\label{fig:wafer_module}
    \protect\subref{wafer_module:exploded_view} 3D-schematic of a \BSSWM\ (dimensions: \SI{50}{\centi\meter} \texttimes{} \SI{50}{\centi\meter} \texttimes{} \SI{15}{\centi\meter}) hosting the wafer~(A) and \num{48} FPGAs~(B).
    The positioning mask~(C) is used to align elastomeric connectors that link the wafer to the large main PCB~(D).
    Support PCBs provide power supply~(E \& F) for the on-wafer circuits as well as access~(G) to analog dynamic variables such as neuron membrane voltages.
	The connectors for inter-wafer and off-wafer/host connectivity (48 \texttimes{} Gigabit-Ethernet) are distributed over all four edges~(H) of the main PCB\@.
    Mechanical stability is provided by an aluminum frame~(I).
    \protect\subref{wafer_module:picture} Photograph of a fully assembled wafer module.
  }
\end{figure}

\Cref{fig:wafer_module} depicts a \BSS{1} wafer module.
The main constituent is a silicon wafer, manufactured in \SI{180}{\nano\meter} Complementary Metal Oxide Semiconductor (CMOS) technology, carrying 384 HICANN chips that are interconnected via an on-wafer bus network.
Each chip hosts up to 512 AdEx neurons and 113k plastic synapses.
48 Xilinx Kintex-7 Field Programmable Gate Arrays (FPGA) provide an I/O interface for configuration, stimulus and recorded data.
The connection between these FPGAs and the control cluster network is established via 1-Gigabit and 10-Gigabit Ethernet, cf.\ \textcite{schmitt2017hwitl}.

\subsection{Performing Experiments}\label{sec:performing_experiments}

Providing access to and experimenting with neuromorphic systems is an active field of research~\cite{rhodes2018spynnaker,rowley2019spinntools,lin2018programming,amir2013cognitive}.
In 2017~\textcite{schuman2017survery} highlighted that:
\begin{quoting}
Supporting software will be a vital component in order for neuromorphic systems to be truly successful and accepted both within and outside the computing community.
\end{quoting}
We second this statement and this paper details our approach on how to tackle the software challenge in the context of \BSS{1} in particular.
However, many of the developed solutions are of general scope and should be applicable to other systems that have similar features and targets.

Different approaches to user interfaces are viable, e.g., the interface for Spikey ---a previous chip developed by the Heidelberg group--- mainly used \codeInline{Python} for configuration and experiment description~\cite{bruederle2009pyhal}.
In contrast, \stackName provides only thin \codeInline{Python} wrappers for all user-facing Application Programming Interfaces (API) while the core software is written in \codeInline{C++}, cf.\ \cref{sec:programming_environment}.
Spikey focused on \codeInline{PyNN} as the experiment description language~\cite{davison2009pynn}, see \cref{sec:pynn}.
As of today, common spiking neural network simulators and a few hardware emulators support \codeInline{PyNN}~\cite{rhodes2018spynnaker,davison2009pynn}.
Further work investigated the typical workflow when porting experiments from pure software simulations to neuromorphic platforms~\cite{bruederle2011comprehensive}.

\BSS{1} has a fairly large parameter space, O(\SI{50}{\mebi\byte} per wafer) of static configuration data, and the analog characteristics of the system require expert knowledge when configuring the system on a low abstraction level.
Thus, an easy-to-use interface and the support of non-expert users are the main objectives of the \BSS{1} software development effort.
In addition to the neuroscientific API, the current \BSS{1} software stack provides access a multitude of interfaces to manipulate the experiment description and system configuration on lower level, cf.\ \cref{sec:halconf,sec:halctrl}.
For example, configuring certain hardware entities in a manual fashion but still being able to rely on the automated process in other areas facilitates the commissioning of new systems.
This approach allows for exposing expert-level manipulation of the complex neuromorphic substrate without giving up all benefits of automation.
Finally, increasing numbers of platform users, the parallel operation of production-type systems as well as commissioning future hardware generations pose challenges for the development and operation methodologies:
platform robustness, experiment reproducibility and, to a lesser extent, the reduction of turnaround times between hardware revisions are essential.

\subsection{Experiment Platform}\label{sec:experiment_platform}

Providing the research community with access to neuromorphic systems has been an ongoing effort pushed by large-scale research projects, such as the Human Brain Project.
In the case at hand, a large-scale neuromorphic system is operated in a multi-user setting.
From a high-performance computing (HPC) perspective, ``fast'' neuromorphic systems resemble spiking neural network accelerators.
To efficiently utilize the available hardware systems requires solutions that are also common in HPC centers:
resource management, time-sharing, fairness, accounting, monitoring and visualization.
However, providing access to external researchers also increases the need for robust operation, experiment reproducibility and support.

\section{Methods and Tools}\label{sec:meths}

Performing well-controlled experiments on the \BSS{1} system is the main task of \stackName.
In broad terms, experiments are defined by \emph{what} and \emph{when}, representing the data and the control flow.
APIs for spiking neural network (SNN) descriptions, e.g., \codeInline{PyNEST} or simulator-agnostic \codeInline{PyNN} typically focus on the initial experiment setup, i.e.\ network topology, model parameters, plasticity rules, recording settings and the stimulus definition.
Similarly, neuromorphic hardware requires an initial configuration that is typically performed before any stimulus is connected to the SNN.

Neuromorphic hardware is also different, as it often needs additional ---more technical--- settings compared to simulators which numerically calculate the time evolution of differential equations as, e.g., described by~\textcite{Einevoll2019scientific}.
In our case, APIs that solely concentrate on aspects of neuron and synapse models, network topology and stimulus are insufficient, especially during commissioning.
Usability for both experts and non-experts is the key feature of the software stack.
The main points of usability are:
\begin{inparaenum}
	\item encapsulation of domain knowledge into software layers;
	\item validity checking of settable parameters;
	\item error reporting and explicit error handling;
	\item consistency in the API layers concepts, and their representation of hardware entities;
	\item availability of tested settings and configuration protocols to the user;
	\item possibility to inject customized behavior at all levels of the software stack.
\end{inparaenum}

The development of the \BrainScaleS{1} platform started already in 2008.
Though many hardware and software components are now over a decade old, updates and improvements are continuously being made.
In this section, we shortly describe the development methodology and foundations for the \stackName.

\subsection{Methodology}

Compared to previous efforts made when developing the software stack for Spikey~\cite{bruederle2009pyhal}, large-scale neuromorphic systems introduce additional complexity.
For example, multi-chip setups require more automation and robustness in all parts of configuration and runtime control.
Hence, more people collaborate on different aspects of the system which, in turn, introduces friction in the development and commissioning process.
When the development of \stackName started, we employed version control, personal interaction and test frameworks.
Within the first four years of development, a chat system~\cite{mattermost} ---also utilized for users support---, continuous integration~\cite{jenkins2015continous} and formalized code review~\cite{gerrit} were added to the development process.
At the time of writing, over \num{10000} changes were submitted for discussion.
We do not adopt a strict development process framework, e.g., like \codeInline{scrum}, however, we do include ourselves within the agile movement.
Weekly meetings provide the scope for structured, long-term development, whereas our chat and code review systems encourage technical discussions of details.
The scenario is similar to CERN-style development where the developers are also to a large extent the users~\cite{brun2012from}.
Over 120 individuals contributed across various projects.
In the following paragraphs, we introduce the key concepts.

\subsubsection{Open Source}
Open Source software is a vital part in almost all fields of research.
If not stated otherwise, the developed libraries and tools are published at \url{https://github.com/electronicvisions} under the \codeInline{LGPL v2.1}~\cite{lgplv21} license.
We also actively report bugs and push features upstream to third-party libraries.

\subsubsection{Software Design}

The long-term software and hardware development roadmaps are aligned to each other.
Weekly software development meetings form the basis of the collaborative development.
Problems and feature requests are discussed as well as medium-term development planning is performed.
If needed, smaller teams are formed to come up with proposals that are then discussed in the plenum.
However, the process is not fully democratic and at the end the maintainers take the final decisions.

\subsubsection{Review}

The BrainScaleS hardware and software development teams adopted an explicit review-based development scheme.
Tracking of the development history and the current state of all components is handled by a set of version-controlled repositories.
Developers propose changes to aspects of this state which are subsequently reviewed by other developers.
At the end, the automated verification of each change and an iterative review process result in a final version which is then applied to the repository and becomes the new current state.
This enables a rolling release scheme.

\subsubsection{Verification}

Based on the ideas of the continuous integration development methodology, the BSS verification methodology consists not only of software tests but also of hardware-based as well as simulation-based tests.
For each proposed change the test result is fed back into the review system.
When changes to software components are applied to the current state, the modified software is automatically deployed.
Nightly tests serve as a measure for hardware platform health.
The same experiment protocol can be used on hardware and in a combined FPGA and digital chip simulation.
The latter is used for pre-tapeout verification during chip development.

\subsubsection{Software Environment}

Reuse of existing software packages reduces development costs but also introduces technical debt in the form of dependencies on external software packages~\cite{cox2019surviving}.
BrainScaleS uses a containerized and explicit software dependency tracking system based on singularity~\cite{kurtzer2017singularity} and spack~\cite{gamblin2015spack}.
Updating the software environment is based on the same review and verification system as before:
developers propose a change to the dependency list,
a testing container image is built and
all tests are executed using this container image.
If the container has build and code review as well as all tests have passed, the proposed change can be applied and the modified container image becomes the new default container image.

\subsection{Foundations} %

\subsubsection{PyNN}\label{sec:pynn}

\codeInline{PyNN}~\cite{davison2009pynn} is a simulator-agnostic domain-specific language for describing spiking neural network models.
Rooted in computational neuroscience, it focuses on the initial network topology, model parameters and plasticity rules, definition of input stimulus and ``recording'' settings.

Matching our goal of an backend-independent experiment description language for spiking neural networks, we adopted \codeInline{PyNN} as our high-level API.
However, BSS-1 is not as flexible as a software simulator.
For example, it only supports a fixed neuron model, limited-resolution synapses and a sparse connectivity matrix.
Transforming a user-defined PyNN experiment into a similar, well-fitting hardware configuration is challenging task.
In particular, it is a matter of neuron and synapse placement, spike routing and model parameter translation.
Due of imperfections of the analog substrate and limited resources, like bandwidth, there will always be differences to the user-defined target.
For a detailed study, see \textcite{bruederle2011comprehensive}.

\subsubsection{Programming Environment}\label{sec:programming_environment}

\codeInline{GNU/Linux} is a flexible and well-supported host environment when developing custom hardware.
In addition, computational neuroscience relies heavily on libraries and tools that are available for \codeInline{*NIX}-like operating systems.
Therefore, we only target \codeInline{Linux}.

All core libraries are written in \codeInline{C++} with the exception of parts of the transport layer that need a tight coupling to the \codeInline{Linux} kernel and are therefore written in \codeInline{C}.

We chose \codeInline{C++} because of several reasons that are not unique to an neuromorphic operating system but apply to requirements of large software suites that have at least a modest need for performance and robustness.
It is a multi-paradigm strongly-typed compiled language which leads to the discovery of many problems at compile time instead of runtime.
\codeInline{C++} supports many low-level manipulations that are essential when directly communicating with a custom hardware system.
For example, in lower software layers the in-memory layout of data structures is required to match the formatting expected by the system.
We always use the latest language standard that typical open-source compilers, e.g., \codeInline{GCC}~\cite{gough2005introduction} and \codeInline{LLVM}~\cite{lattner2004llvm} support.

\subsubsection{Python Wrapping}

In experimental usage settings, scripting languages offer large advantages compared to compiled languages.
For example, the read–eval–print loop (REPL) allows for iterative testing of the hardware and also for exploration of the software itself.
Integration with the broad \codeInline{Python} ecosystem of scientific libraries, e.g., \codeInline{numpy}~\cite{walt2011numpy} or \codeInline{matplotlib}~\cite{Matplotlib2007}, is an advantage in scientific efficiency.
Therefore, we support \codeInline{Python} in addition to \codeInline{C++}.
To link the \codeInline{Python} and the \codeInline{C++} world, we adopted a fully automated wrapper code generation scheme based on \codeInline{py++} and \codeInline{pygccxml}~\cite{pygccxml}.
The generated wrapper code uses \codeInline{boost::python}~\cite{boostpython_1710_homepage}.
Customizations of the wrapping process have been collected in a library. %

\subsubsection{Serialization}\label{sec:serialization}

Serialization describes the process of transforming in-memory data structures or object states into a store and loadable format.
This format can be written to, e.g., disk and loaded to restore the in-memory data structures at a later point in time.
Together with a framework for remote procedure calls, such as \codeInline{RCF}~\cite{RCF}, this allows for inter-process communication of higher-level data structures.
Though \codeInline{C++} does not offer built-in support for serialization, it can be made available through several third party libraries.
For \stackName{}, we decided to use \codeInline{boost::serialization}.

\begin{listing}[htbp]
  \caption{Example for \codeInline{boost} serialization.}
  \begin{minted}{c++}
  class Spike
  {
  // ...
  private:
    template<typename Archiver>
    void serialize(Archiver& ar,
                   unsigned int const /*version*/)
    {
      using boost::serialization::make_nvp;
      ar & make_nvp("addr", m_addr)
         & make_nvp("time", m_time);
    }

    addr_type m_addr;
    time_type m_time;
  };
  \end{minted}
  \label{lst:boost_serialization}
\end{listing}

\Cref{lst:boost_serialization} exemplifies the serialization of the two member variables of the \codeInline{Spike} class.
More complex serialization functions are needed, e.g., when references and pointers are involved or different versions should be considered to support long-term compatibility for pre-existing data sets.
However, \codeInline{boost::serialization} has excellent support for all these scenarios.
The on-disk format ranges from binary to text-based, such as \codeInline{JSON} (custom extension) and \codeInline{XML}.

\subsubsection{Utility Libraries}

\paragraph{Ranged enumeration types}
In \codeInline{C++} numeric types do not have built-in support for range checks.
Yet, it is beneficial to have such concepts, since over and under runs can be a threat to both correctness and security.
The \codeInline{rant}~\cite{githubrant} library provides ranged integers and provides compile-time as well as runtime checking of ranges.
In case of compile-time statements, violations produce compile errors;
runtime errors raise exceptions.
Checks are implemented to be lightweight enough to be included in production code.
If required, ranged types can be replaced by their native counterparts via a compile flag to get rid of any remaining overhead.
Ranged integers are heavily used for the coordinates described in the next section.
\Cref{lst:rant} demonstrates the use on the example of a ranged integer type.

\begin{listing}[htbp]
  \caption{Example for a ranged type.}
  \begin{minted}{c++}
  rant::integral_range<int, 0, 5> ranged_integer;

  // e.g. 6 => throws
  ranged_integer = get_large_int();

  // fails to compile (constexpr)
  ranged_integer = -1;

  // works
  ranged_integer = 3;
  \end{minted}
  \label{lst:rant}
\end{listing}

\paragraph{\codeInline{Python}-style \codeInline{C++} convenience library}

\codeInline{pythonic++}~\cite{githubpythonic} brings some \codeInline{Python}-style programming to \codeInline{C++} for ease and more expressive code.
\Cref{lst:pythonic_enumerate} exemplifies this idea on the enumeration during the iterations over a vector.

\begin{listing}[htbp]
  \caption{The \codeInline{pythonic::enumerate} function can be used to count the iterations over an \codeInline{STL} conform container.}
  \begin{minted}{c++}
  using namespace pythonic;

  typedef std::vector<int> vec;

  for (auto v : enumerate(vec{0, -1337, 42}))
  {
    std::cout << v.first << " " << v.second << '\n';
  }
  \end{minted}
  \label{lst:pythonic_enumerate}
\end{listing}

\paragraph{Bit Manipulation Library}

\codeInline{bitter}~\cite{githubbitter} provides a common interface for bit operations on integral types as well as \codeInline{std::bitset}.
Operations like reversal or cropping to ranges are implemented.

\section{Implementation}\label{sec:impl}

The \stackName{} consists of several software component categories which are described in the following sections, see \cref{fig:swstack} for an overview.
At the end, the user is enabled to describe and execute a neuromorphic experiment without detailed knowledge of the underlying parts.

\begin{figure}[tbp]
  \hspace{0.5cm} %
  \begin{tikzpicture}
    \def\centerwidth{.75cm}
    \def\layerwidth{.75cm}
    \node[circle, fill=gray!50, thick, minimum size=2*\centerwidth, inner sep=0] {BSS-1};
    \def\labels{FPGAs and Communication Layer}
    \StrCount{\labels}{,}[\labellen] %
    \pgfmathsetmacro\labellen{\labellen+1} %
    \pgfmathsetmacro\slice{360/\labellen}
    \def\sliceoffset{270+\slice} %
    \foreach \label [count=\i from 0] in \labels {%
      \pgfmathsetmacro\startangle{180 - \slice*\i - \sliceoffset}
      \drawsector[thick, fill=green!30]{\centerwidth}{\centerwidth}{\startangle}{\slice}{\label}
    }
    \def\labels{Map and Route, Hardware Abstraction, Bit Formatting}
    \StrCount{\labels}{,}[\labellen]
    \pgfmathsetmacro\labellen{\labellen+1}
    \pgfmathsetmacro\slice{360/\labellen}
    \def\sliceoffset{90+\slice} %
    \foreach \label [count=\i from 0] in \labels {%
      \pgfmathsetmacro\startangle{180 - \slice*\i - \sliceoffset}
      \drawsector[thick, fill=blue!20]{2*\centerwidth}{\centerwidth}{\startangle}{\slice}{\label}
    }
    \def\labels{PyNN Implementation, Calibration, Blacklisting}
    \StrCount{\labels}{,}[\labellen]
    \pgfmathsetmacro\labellen{\labellen+1}
    \pgfmathsetmacro\slice{360/\labellen}
    \def\sliceoffset{30}
    \foreach \label [count=\i from 0] in \labels {%
      \pgfmathsetmacro\startangle{180 - \slice*\i - \sliceoffset}
      \drawsector[thick, fill=yellow!20]{3*\centerwidth}{\centerwidth}{\startangle}{\slice}{\label}
    }
    \def\labels{Sampling, Time-to-first Spike, Rate-based Deep Learning}
    \StrCount{\labels}{,}[\labellen]
    \pgfmathsetmacro\labellen{\labellen+1}
    \pgfmathsetmacro\slice{360/\labellen}
    \def\sliceoffset{60}
    \foreach \label [count=\i from 0] in \labels {%
      \pgfmathsetmacro\startangle{270 - \slice*\i - \sliceoffset}
      \drawsector[thick, fill=red!20]{4*\centerwidth}{\centerwidth}{\startangle}{\slice}{\label}
    }
  \end{tikzpicture}
  \caption{\label{fig:swstack}The software stack as layers of abstraction.
	The core is the neuromorphic hardware \BSS{1}.
	It is followed by physical and software communication layers, i.e.\ the FPGAs and communication layer,
    followed by hardware abstraction and core functionality like mapping and routing.
	Next comes expert-level software used for blacklisting and calibration and the layer providing the \codeInline{PyNN} abstraction.
    The last layer are the user-level applications and experiments.
  }
\end{figure}
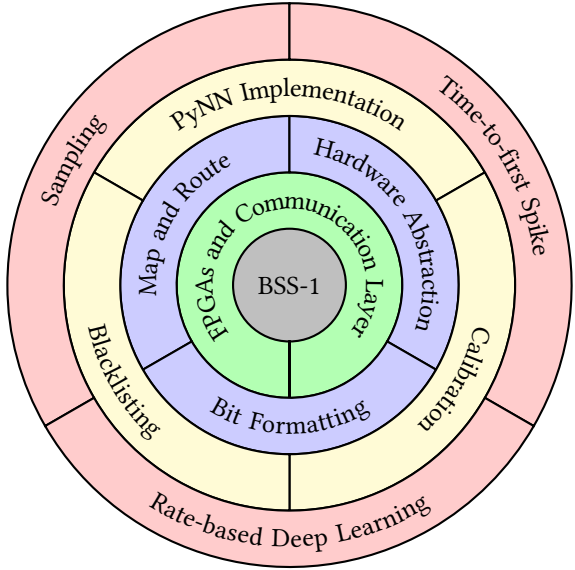

\subsection{Configuration}\label{sec:halconf}

The correct configuration of any hardware is a non-trivial task, e.g., one needs to ensure to write the correctly formatted command to the correct memory location, while taking into account other subtleties as, e.g., configuration order.
In addition, there can be a mismatch between the addressing of individual circuit instances and their physical or logical placement.
Another aspect is configuration timing as some entities require settling times that have to be taken into account.
To address the issues raised, the \stackName{} comprises of several libraries that allow for an easy and correct configuration and control flow.

\subsubsection{Coordinate System}\label{par_coord}
In natural sciences, the proper choice of a coordinate system often strongly contributes to a simple and clean solution for a problem.
We argue that this applies also to the usage of hardware in general and in particular to neuromorphic computing.
The myriad of components on a wafer lead to a configuration space in doe order of \SI{50}{\mebi\byte} \cite{bruederle2011comprehensive}.
The memory for 4-bit weight and 4-bit address filter of the 384 chips \texttimes{} 113k synapses per chip alone amounts to \SI{41}{\mebi\byte} per wafer.
One needs a representation of those components in software~\cite{githubhalco}.

Many symmetries in chip layout combined with wafer-scale integration naturally lead to abstraction on different scales.
\Cref{fig:coordinates} gives an overview of a \BSS{1} wafer and the structure of its components.
Framed in blue one can see the layout of a single HICANN chip and its high degree of self-similarity.
In the background one silicon wafer containing 384 of those chips is shown.
This translational symmetry is reflected in a hierarchical structure of the coordinates.
For each component we define a coordinate with the smallest granularity which then can be combined to define a higher hierarchical layer.
We will illustrate this exemplarily with the coordinate for neuron circuits.
First, we represent one neuron circuit on a chip:
\codeInline{NeuronOnHICANN}.
This can then be combined with \codeInline{HICANNOnWafer} resulting in \codeInline{NeuronOnWafer} representing a specific neuron circuit on a wafer.
Finally, \codeInline{NeuronGlobal} can be composed from \codeInline{NeuronOnWafer} and \codeInline{Wafer} to uniquely identify one neuron circuit in the whole \BSS{1} system.
It is also possible to cast down to lower levels of representation, e.g., \codeInline{NeuronOnWafer::toNeuronOnHICANN()}.
Besides ``lateral'' conversions between hierarchical layers it is also possible to translate ``horizontally'' among coordinates on the same level.
For example \codeInline{SynapseOnHICANN::toNeuronOnHICANN()} yields the matching \codeInline{Neuron} of a \codeInline{Synapse}.
See \cref{lst:coord_conv} for additional examples.

Another important feature is the possibility to create two dimensional grids that also have a notion of orientation, e.g., north and south.
\codeInline{SynapseOnHICANN} for example is structured in a grid of neurons per chip hemisphere and synapses per neuron. %
Grid coordinates also provide enumeration which is done in row-major order as shown in orange in \cref{fig:coordinates}.
Enumeration enables iteration of all coordinates which is supported in both, \codeInline{C++} and \codeInline{Python}, cf.\cref{lst:coord_iter_cpp} and \cref{lst:coord_iter_cpp}.
A string serialization exists that serves as both, convenient short format for logging and for argument parsing, cf.~\cref{sec:redman}.
An example for this functionality can be found in \cref{lst:coord_short_format}.
The consistence of this hierarchical structure is essential for a descriptive, reliable and maintainable low-level code base.
About 80 distinct coordinate types are used to describe elements of a wafer module.

\begin{listing}[htbp]
  \caption{Example coordinate conversion.}
  \begin{minted}{python}
  nrn = NeuronOnWafer(NeuronOnHICANN(Enum(5)), HICANNOnWafer(Enum(5)))
  nrn.toNeuronOnHICANN()
  nrn.toHICANNOnWafer()
  \end{minted}
  \label{lst:coord_conv}

  \caption{Example coordinate iteration in \codeInline{C++}.}
  \begin{minted}{c++}
  for(auto nrn : iter_all<NeuronOnHICANN>()) {
    std::cout << nrn << '\n';
  }
  \end{minted}
  \label{lst:coord_iter_cpp}

  \caption{Example coordinate iteration in \codeInline{Python}.}
  \begin{minted}{python}
  for nrn in iter_all(NeuronOnHICANN):
      print(nrn)
  \end{minted}
  \label{lst:coord_iter_py}

  \caption{Example coordinate short formatting.}
  \begin{minted}{python}
  h = HICANNGlobal(HICANNOnWafer(Enum(5)), Wafer(6))
  print(short_format(h))
  # W006H005
  print(from_string("W3"))
  # Wafer(3)
  \end{minted}
  \label{lst:coord_short_format}
\end{listing}

\begin{figure}[tbp]
    \includegraphics[width=\columnwidth]{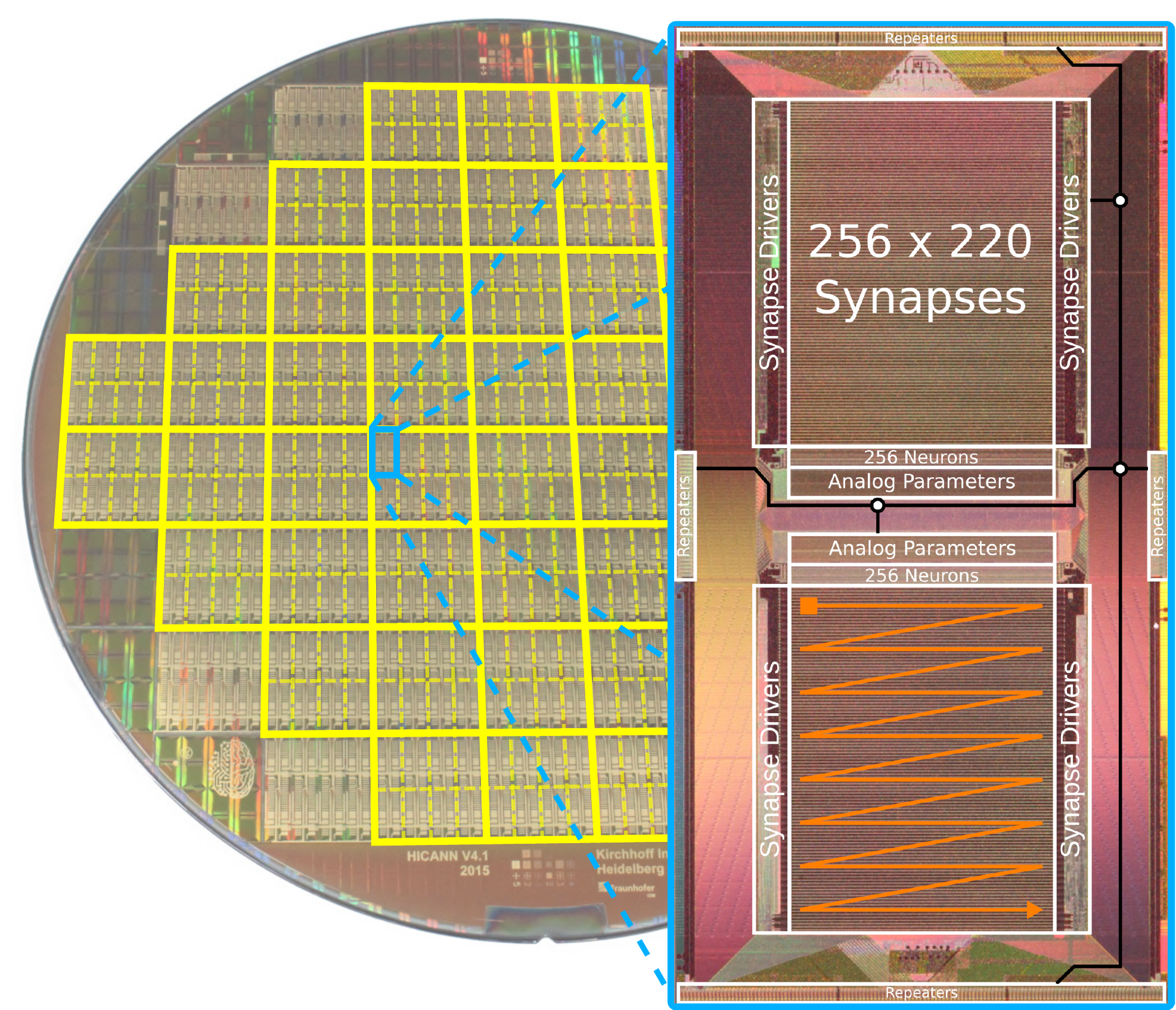}
    \caption{\label{fig:coordinates}
    Chip structure and coordinate system of a \BSS{1} wafer:
    the background shows a silicon wafer with highlighted structures of chips grouped in units of 4-by-2. %
	The zoom-in shows one single HICANN chip layout.
    Framed in white are various component categories.
    Black lines illustrate the structure of intra- and inter-chip event buses.
    The row-major ordering scheme of a two-dimensional coordinate is shown in orange over a synapse array.
    }
\end{figure}

\subsubsection{Bit Formatting}

Typically, to configure a hardware entity a pair of an address and its content is needed.
Addresses are used to identity writable and/or readable memory locations.
The formatting of the content may depend on various aspects, e.g., the entity's physical location on the chip.
Therefore, functionality to address hardware entities and to format bits is an essential part of the software~\cite{githubhalbe}.
We use coordinates, \cref{par_coord}, to logically represent addresses.
The content is represented as data structures encapsulating functionality of the underlying entity, cf.~\cref{lst:halbe_synapse_driver}.

\begin{listing}[htbp]
  \caption{Excerpt of the data structure \codeInline{SynapseDriver}.}
  \begin{minted}{c++}
  struct SynapseDriver {
  public:
      bool is_enabled() const { return enable; }
      ...
  private:
      bool enable = false
      ...
  };
  \end{minted}
  \label{lst:halbe_synapse_driver}

  \caption{Example for a pair of getter and setter functions of the stateless hardware abstraction layer.}
  \begin{minted}{c++}
  void set_synapse_driver(
      Handle::HICANN& handle,
      SynapseDriverOnHICANN const& sdoh,
      SynapseDriver const& driver);

  SynapseDriver get_synapse_driver(
      Handle::HICANN& handle,
      SynapseDriverOnHICANN const& sdoh);
  \end{minted}
  \label{lst:halbe_getter_setter}
\end{listing}

During development, for commission, expert use and debugging, an iterative \& interactive usage is facilitated by \codeInline{Python} bindings for the lower-level configuration functions.
This allows to, e.g., change parts of the configuration ---also out of order w.r.t.\ the canonical flow--- and directly observe the effects.

\Cref{lst:halbe_getter_setter} gives an example for a pair of getter and setter functions.
The \emph{handle} represents the backend, either accessing the hardware, the simulation backend described in \cref{sec:ess}, or other debug facilities.
The coordinate \codeInline{SynapseDriverOnHICANN} specifies for which synapse driver the settings in the \codeInline{SynapseDriver} container \codeInline{driver} should be applied.
The getter function is the same in reverse.
Here, only the handle and the coordinate are passed.
The bits read back from the hardware are decoded into and returned as a \codeInline{SynapseDriver} object.

\subsubsection{High-level Configuration}

The core principal of the configuration of the neuromorphic hardware is that the user first specifies the desired state to which then the hardware is configured to~\cite{githubsthal}.
Then, the hardware is configured to this state.
To facilitate this user-driven configuration, all configurable settings have functional names, e.g., the neuron configuration.
By this, a viable level of ``self-documentation'' is achieved.
The user-facing configuration does not necessarily reflect the exact granularity in which the hardware can be configured, however, it does reflect, as stated above, an achievable final state which then also allows validity checks.

\begin{listing}[htbp]
  \caption{Example for the stateful hardware abstraction layer.}
  \begin{minted}{c++}
  sthal::Wafer wafer;
  auto& hicann = wafer[HICANNOnWafer(Enum(5))];
  hicann.synapses[SynapseOnHICANN(Enum(123))].weight = SynapseWeight(3);
  \end{minted}
  \label{lst:sthal_usage}

  \caption{Example for per-FPGA parallelism via \codeInline{OpenMP}.}
  \begin{minted}{c++}
    #pragma omp parallel for schedule(dynamic)
    for (size_t fpga_enum = 0; fpga_enum < FPGAOnWafer::end; ++fpga_enum) {
      ...
    }
  \end{minted}
  \label{lst:fpga_parallel_openmp}
\end{listing}

\Cref{lst:sthal_usage} demonstrates how to set the weight of a single synapse.
The needed objects and bookkeeping structures are created on the fly.
Also, checks on the availability database, see \cref{sec:redman}, are performed and raise exceptions if the requested resources are not available.

The configuration is carried out with the maximum parallelism supported by the system, e.g., on a per-FPGA-basis with the help of \codeInline{OpenMP}, see \cref{lst:fpga_parallel_openmp}.

\subsection{Control}\label{sec:halctrl}

\subsubsection{Experiment Control Flow}\label{sec:ctrlflow}

The \BSS{1} platform supports two distinct operation modes, both relying on FPGAs for data I/O and for control flow.
Figure \ref{fig:control_flow} illustrates the control flow for the primarily used mode, the \emph{batch} mode, which suits independent pre-defined experiments.

In either case, the first step is configuring the neuromorphic hardware.
Many configuration register accesses on the chip use a non-blocking access scheme requiring correct timing.
This is implemented by inserting wait instructions between configuration commands;
the time intervals use a static worst-case timing model.
It is also important to configure hardware entities in a valid order.
This is especially relevant when entities are configured in parallel, e.g., per FPGA\@.
Synchronization barriers must then be added to the configuration flow so that only when all entities have reached a certain stage, configuration is continued.
Another point to take care of is enabling triggers for, e.g., the recording of analog membrane traces which are supposed to start with the experiment;
it is the point in time when the stimulus begins and recording of spike events is enabled.

Timed spike event release as well es recording is handled by 48 FPGAs on each wafer module.
Each FPGA has accesses to \SI{1.25}{\gibi\byte} of DRAM providing buffer memory for, e.g., input stimulus and recorded data.
In case of the batch mode, the complete input to the network is predefined on the host, then sent to the FPGA and released upon experiment start.
Simultaneously, spikes generated by the neuromorphic chips are recorded.
At the end of the experiment, the recorded spikes and analog membrane traces are sent back to host.
One typical application of this mode are deep neural networks where synaptic weights are optimized by an offline learning algorithm.
If the hardware is allocated for a longer time period, the experiment framework also supports selectable automatic differential configuration reducing configuration overhead in later iterations.

In the other so called \emph{hybrid} operation mode parts of the network or a virtual environment are simulated on the control cluster and interact with the spiking neural network running on \BSS{1}.
This mode of operation is also known as \emph{real-time closed-loop}.
Control flow differs compared to the aforementioned mode as spike events sent by the host are not pre-buffered and timed by the FPGA but instead, they are directly injected into the chip upon arrival.
Vice versa, emitted spikes from the network are directly sent to host and reacted upon.
The challenge is to match the acceleration factor in both realms and keep the latency of the network communication as low as possible, see \cref{par_tp_layer}.

\begin{figure}[tbp]
	\includegraphics[width=\columnwidth]{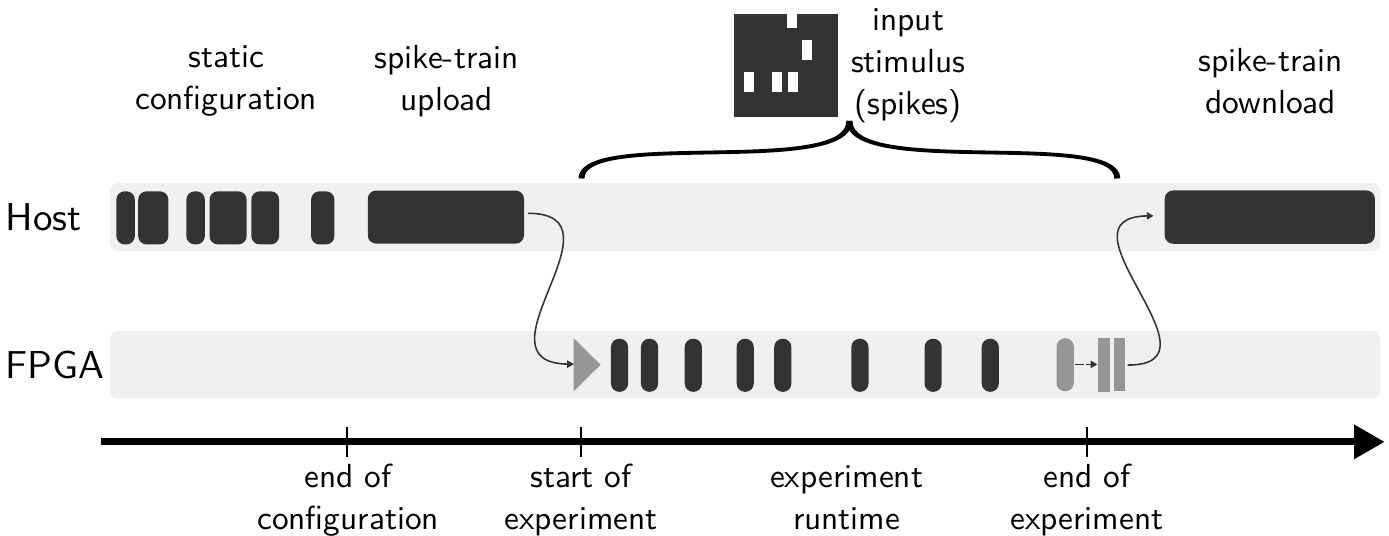}
	\caption{Control flow of a typical experiment in batch mode.
             Black boxes indicate activity of host and FPGA during the different steps.
             \label{fig:control_flow}
    }
\end{figure}

\subsubsection{Communication Layer}\label{par_tp_layer}

There are two main categories of data which need to be transferred between the neuromorphic hardware and conventional compute nodes.
On the one hand there is configuration data, e.g., neuron parameters, network topology, and on the other hand the activity of the network, i.e.\ spike events.
Due to the accelerated operation of the \BSS{1} system typical neuron activities of O(\SI{100}{\hertz}) result in on-wafer event rates exceeding Tera-Events per second.
This demands high-throughput data exchange between \BSS{1} and cluster control nodes.
Ethernet was chosen as conventional data network equipment is readily available and, at the time of writing, commercial hardware supports bandwidths of up to \SI{100}{\gibi\byte\per\second}.
On \BSS{1}, the external connectivity is provided by 48 1-Gigabit Ethernet links.
However, this bandwidth is still not sufficient to completely monitor the aforementioned on-wafer activity.
The FPGAs provide an additional buffer stage for input and output data, but filtering and selecting in- and outputs is still needed.
In the case of a deep neural network this would for example simply be the in- and output layers.

Furthermore, transfer of data, especially configuration data, needs to be robust.
For Ethernet-based communication the Transmission Control Protocol (TCP) on top of the Internet Protocol (IP) is most commonly used as a reliable secure transport layer protocol.
At the time of development there where no open source FPGA implementations of TCP available and even now available solutions are very resource demanding \cite{ruiz2019opentcp}.
Hence, the \BSS{1} FPGAs implement a custom sliding-window protocol with an automatic resend mechanism (ARQ) on top of the unreliable User Datagram Protocol (UDP) over IP.
The software implementation has been open-sourced in the past~\cite{githubhostarq}.
Additional features analogous to congestion control, like roundtrip-time estimation as well as the slow start algorithm have been implemented in both, software and hardware.

However, the hybrid operation mode, cf.~\cref{sec:ctrlflow}, demands low-latency and low-jitter transport of spike events;
configuration data is still transmitted via the reliable custom transport layer protocol but spike events are transferred best-effort facilitated by memory-mapped zero-copy receive and transmit ring buffers based on \codeInline{PACKET\_MMAP}~\cite{packet_man}.

Additional measures like setting CPU core affinity are taken to reduce jitter to a minimum on the host side.

\subsubsection{Hardware Simulation}\label{sec:ess}
The so-called ``executable system specification''~\cite{githubess} (ESS) is a hardware simulator
of the \BSS{1} system implemented in \codeInline{C++\slash{}SystemC}.
It contains behavioral, timing-accurate models of the digital components and
functional models of the analog neural components, e.g., the hardware neurons
are numerically simulated AdEx neurons.
Offering the same configuration interface as the real hardware and being fully
executable, the ESS has been essential for the hardware-software co-design
\cite{bruederle2011comprehensive} and still serves as a validation tool for the
software stack, especially for the mapping, configuration and experiment
execution steps.
In addition, the ESS allows to evaluate the effect of \BSS{1} design-specific
constraints (e.g., limited stimulation and recording bandwidth, spike time
jitter, reduced parameter resolution) on neuromorphic experiments in isolation
from distortions due to the mismatch of the mixed-signal circuits.
For a detailed study see \textcite{petrovici2014characterization}.

\subsection{Conditions Support}

Wafer-scale hardware operates under the assumption that individual components can be switched off and circumvented.
In addition, the analog nature makes it necessary to, at least, apply a working point calibration.
For this, conditions support libraries are put in place and described below.

\subsubsection{Availability Database}\label{sec:redman}

Errors during the manufacturing process and the assembly of the wafer lead to varying conditions of individual components.
Moreover, modifying hardware parameters may lead to a change in the response of these components.
Disregarded, they might either distort simulation results or make the execution of experiments impossible in the first place.
Consequently, it is mandatory to be aware of the state of the components and handle it dynamically.
Therefore, the availability database was developed~\cite{githubredman}.

Combined with digital tests, cf.\ \cref{sec:blacklisting}, this allows for storing and handling of the used components.
It is implemented in \codeInline{C++} and uses \codeInline{XML} with \codeInline{boost::serialization} as the storage backend.
Based on the coordinate system it stores a sparse representation of the flagged components without the notion of reasons as a whitelist or a blacklist.
By this, the natural hierarchy of the system is mapped to the database. Thus, e.g., \codeInline{HICANNOnWafer} flags the full chip and \codeInline{NeuronOnHICANN} flags only a single neuron circuit of the chip.
Subsequently, using the database, other parts of the software can simply avoid the flagged components.

This also allows for the second use case of the availability database.
Components can be marked artificially as not available to manipulate the hardware resources of an experiment without the requirement of an additional interface.
\codeInline{Python} bindings allow to construct convenience tools like a command line interface, cf.\ \cref{lst:redman_cli}.
As a result, a per experiment set of usable components can be generated and adapted dynamically.

\begin{listing}[htbp]
  \caption{Availability database command line interface.}
  \begin{minted}{shell}
  redman_cli.py . W33H0 has neuron 0
  # True

  redman_cli.py . W33H0 disable neuron 1
  \end{minted}
  \label{lst:redman_cli}
\end{listing}

\subsubsection{Parameter Translation and Calibration Database}

Microelectronics' manufacture deals with non-uniformities in the circuits produced across a silicon wafer.
These transistor mismatches result in varying response from neuron to neuron circuit.
To compensate, a calibration framework~\cite{githubcalibtic} maps high-level parameters to the hardware parameter space, homogenizing the response of neuron circuits.

First, the biological units are converted to the hardware compatible range.
For time constants, the acceleration factor $\alpha=\num{1000}\ldots \num{10000}$ is taken into account:
$$\tau_{\text{hardware}} = \alpha\cdot\tau_{\text{biology}}.$$
Voltages also need to be scaled by $s$ and shifted by $o$, respectively:
$$V_{\text{hardware}} = s\cdot V_{\text{biology}} + o,$$
with typical values $s=10$ and $o=\SI{1.2}{\volt}$
Similar conversions are needed for synaptic weights.

Now that the desired hardware values are known in physical units, the conversion to the digital domain can happen.
This step does two things.
The translation from physical units to digital units while at the same time taking into account variations from circuit to circuit, i.e.\ it applies calibration data.
For this, the calibration database allows to store parameters for a set of pre-defined functions, e.g., polynomials.
In addition, the transformation classes provide a numerical function inversion.

Also, the input values can be checked to lie within a given range of validity.
The returned value can then be either clipped, an exception can be thrown or the validity range can be ignored.

\Cref{lst:calibtic} demonstrates the usage of the library on the example of a linear function.

\begin{listing}[htbp]
  \caption{Example for a linear calibration function.}
  \begin{minted}{c++}
  // linear transformation from, e.g., 0 - 1.8 V to 0 - 1023 DAC
  Polynomial linear({0.0, 1023./1.8}, 0.0, 1.8);
  linear.apply(0.9);
  // 511.5
  linear.reverseApply(256);
  // 0.450
  linear.apply(2); // defaults to clipping
  // 1023
  \end{minted}
  \label{lst:calibtic}
\end{listing}

\subsection{Network Description}

\subsubsection{\codeInline{PyNN} Interface}

We implement the \codeInline{PyNN}-API as a thin \codeInline{C++} library for which \codeInline{Python} bindings are generated~\cite{githubpyhmf}.
Compared to a \codeInline{Python}-based implementation, this allows for a memory-efficient handling of larger data sets such as weight matrices of large neural networks, stimulus or recorded data.
For the user, however, it appears like any other \codeInline{PyNN} implementation, e.g., \codeInline{PyNN.nest}.
Internally, it translates from the \codeInline{PyNN}'s imperative experiment description to an object-oriented description in the underlying \codeInline{C++} layer.
The individual elements, e.g., populations and projections, are similar in their structure to \codeInline{PyNN}.
However, if found necessary, we restructure the data to our liking as it is decoupled from the user-facing API.
Also, pure \codeInline{C++} usage is supported in a structured way.

\subsubsection{Map \& Route}

Mapping and routing of neural networks described in \codeInline{PyNN} to the neuromorphic hardware is a non-trivial task.
The complexity and scope of the problem is similar to the synthesis of FPGA bitfiles.
Therefore, the process is only described briefly.
The full implementation can be found at~\cite{githubmarocco}.
Also, the map \& route implementation undergoes substantial changes as new features and improvements are being developed.

In its simplest form, we implement a greedy strategy without back tracking.
First, we place neurons from \codeInline{PyNN} populations to hardware neuron circuits.
Hereby, different neurons may be represented by a different number of neuron circuits on the hardware.
Insertion points for spike input from the FPGAs are placed as well.
The user has the option to constrain the automatic placement of neurons and spike sources.
User parametrization is facilitated by a custom class that works aside from \codeInline{PyNN}.
\Cref{lst:pymarocco} shows how a user can restrict the placement of a population to a certain HICANN or to a list of allowed options.

\begin{listing}[htbp]
  \caption{Example for constraining placement.}
  \begin{minted}{python}
  pop = pynn.Population(...)
  marocco.manual_placement.on_hicann(pop, HICANNOnWafer(Enum(42)))

  pop2 = pynn.Population(...)
  marocco.manual_placement.on_hicann(pop2, [HICANNOnWafer(Enum(23)), HICANNOnWafer(Enum(24))])
  \end{minted}
  \label{lst:pymarocco}

  \caption{Example for querying a mapping result, where the hardware neurons corresponding to \codeInline{PyNN} neurons are retrieved.}
  \begin{minted}{python}
  pop = pynn.Population(5, ...)
  for pynn_neuron in enumerate(pop):
    items = runtime.results().placement.find(pynn_neuron)
    for item in items:
       for hardware_neuron in item.logical_neuron():
          ...
  \end{minted}
  \label{lst:pymarocco_reverse_mapping}
\end{listing}

After the placement of neurons, the \codeInline{PyNN} projections are transformed into synapses and on-chip routes on the hardware.
This is the most time-consuming step as several hardware constraints must be taken into account, e.g., the limited number of allowed switches per route.

\Cref{lst:pymarocco_reverse_mapping} shows how to retrieve information on the allocated hardware after the placement:
the hardware neuron circuits are looked up for all neurons of a \codeInline{PyNN} population.
This is useful for, e.g., directly manipulating low-level hardware parameters.
The link between \codeInline{PyNN} and the result of the map \& route step is stored into an intermediate representation format, cf.\ \cref{fig:pynn-to-hwcfg}.

\begin{figure}[tbp]
	\includegraphics[width=\columnwidth]{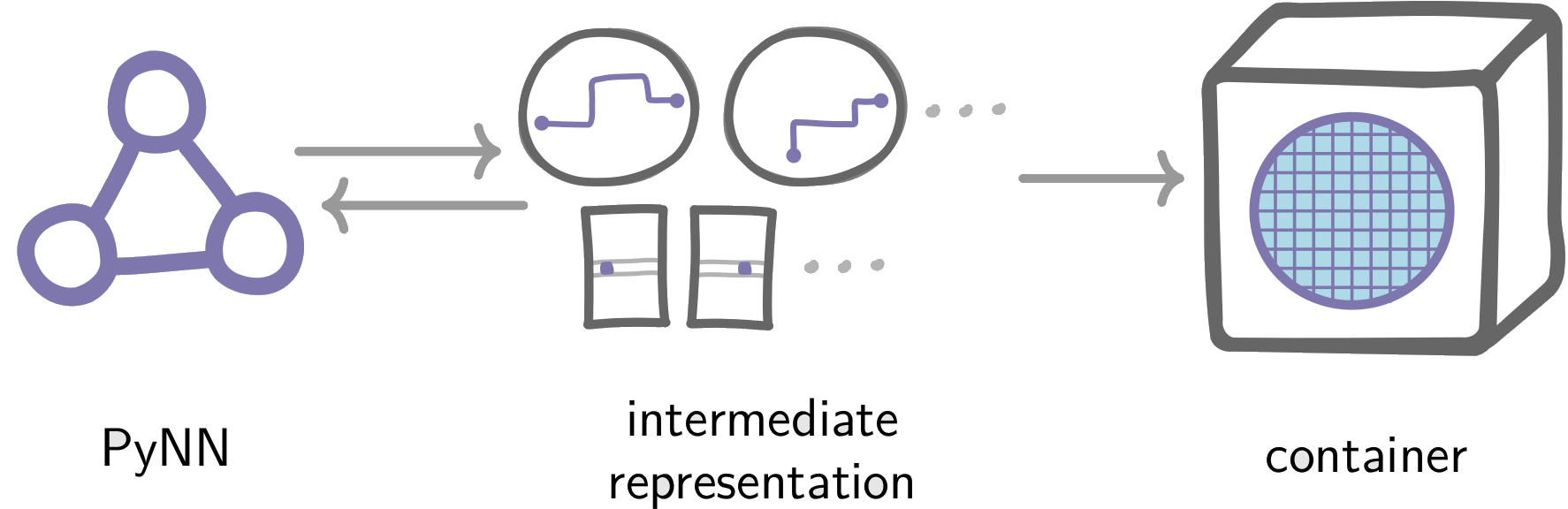}
	\caption{\label{fig:pynn-to-hwcfg}The transformation of \codeInline{PyNN} to a hardware configuration (container) makes use of intermediate representations (IR).
      The IR also links \codeInline{PyNN} and hardware entities and allows for look-ups in both directions.
    }
\end{figure}

The on-chip bus network is represented as a graph using the \codeInline{boost::graph}~\cite{boostgraph_1710_homepage} library, where bus lines are vertices and switches are edges.
During the creation of the graph, hardware availability data is already taken into account, i.e.\ hardware components that should not be used are not included in the graph representation.
On-wafer routes can be found by custom traversal algorithms of the graph or by using graph search algorithms like Dijkstra.

\begin{figure}[tbp]
  \centering
  \includegraphics[width=0.8\linewidth]{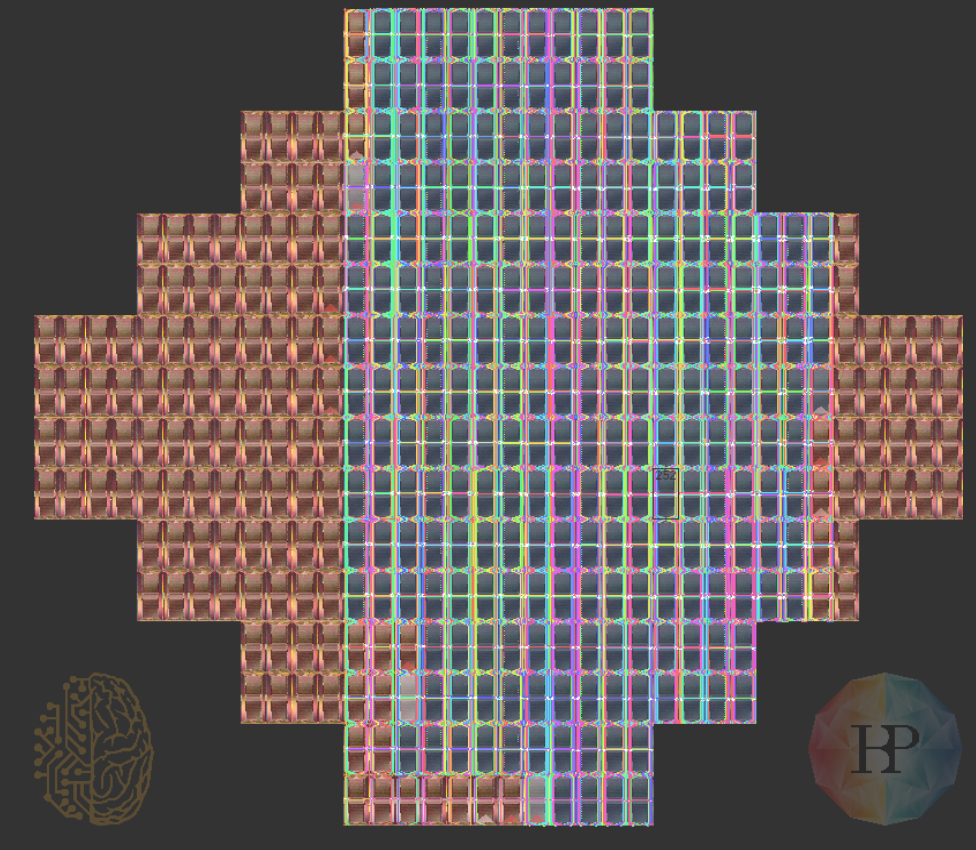}
  \caption{\label{fig:column_mapping_visu}Screenshot of the web-based visualization.
    Chips are colored with increasing opacity proportional to the number of placed neurons.
    On-chip routes are also colored and can be click-selected to reveal more details.
  }
\end{figure}

Being able to have a visual representation of the found hardware configuration is important for both, debugging and understanding possible improvements of manual or automatic placement.
For this, a web-based visualization has been developed.
Based on previous efforts to build visualization tools, the main requirement is to not replicate any code paths that are already part of the software stack.
Another requirement is the possibility to run the tool locally and standalone, i.e.\ without the need for a server and the availability of the full software stack.
This is possible by transpiling parts of the \stackName \codeInline{C++} libraries to \codeInline{JavaScript}, including classes representing the map \& route intermediate representation and its serialization implementation.
Now, only the transpiled \codeInline{JavaScript} libraries and the output of the mapping must be at hand.
The top-level code is written in \codeInline{TypeScript}~\cite{typescript}.

We rely on \codeInline{Pixi.JS}~\cite{PixiJS} for a fast 2D graphics engine supporting WebGL\@.
It is capable of rendering large networks with many details of the hardware configuration.
The tool offers different levels of details where, e.g., by zooming in all used synapses and neurons become visible.
An example is shown in \cref{fig:column_mapping_visu}.

\section{Operation}\label{sec:operation}

\subsection{Resource Management and System Access}

BrainScaleS neuromorphic platform resources are time-shared and partitioned between multiple experiments and/or users.
In contrast to typical digital systems, analog neuromorphic hardware substrates are not homogeneous.
Users need to be able to request specific hardware instances when running experiments.
We use \codeInline{SLURM}~\cite{yoo2003slurm} ---a HPC job scheduler--- to handle resource requests for hardware components.

\codeInline{SLURM} was extended utilizing its plugin API to handle various requirements related to inhomogeneous hardware resources.
Being a mixed-signal neuromorphic system, individual BrainScaleS systems behave slightly differently which is why experimenters need a way to explicitly specify individual resource instances.
The coordinate system described in \cref{par_coord} is used to provide a familiar interface to the user.
Different hardware components have varying degrees of granularity that can overlap and have interdependencies.
We allocate the smallest needed subset of resources inferred from the user request.
In principle, the Ethernet-based communication described in \cref{par_tp_layer} allows access to each FPGA from any conventional compute node in the same network.
To prevent accidental clashes between concurrent experiment runs we separate individual \BSS{1} modules into IPv4 subnets and deny access based on default firewall settings.
When a user specifies a hardware resource for a \codeInline{SLURM} job a firewall rule to accept traffic is automatically added during job runtime. %
Experiment software also compares its own resources with the allocated \codeInline{SLURM} resources to detect possible mismatch.

On top of the direct access to the system as explained above, we provide access via HBP's collab infrastructure~\cite{amunts2016hbp}.
Jobs are fetched from the HBP neuromorphic platform queuing service with the help of~\cite{hbp-neuromorphic-client}
and passed on to \codeInline{SLURM}.
Every few seconds, the job states of our scheduler and the upstream queue manager are synchronized.

\subsection{Monitoring}

Managing a large complex hardware system is unfeasible without extensive monitoring, as malfunction of any individual component can be fatal for operation.
Monitoring can generally be split into three steps: aggregation, storage and visualization.
Likewise there are two different types of data to be gathered, time-series data (e.g., voltage, temperature) and event data (e.g., powering off components).

\begin{figure}[tbp]
	\includegraphics[width=.5\textwidth]{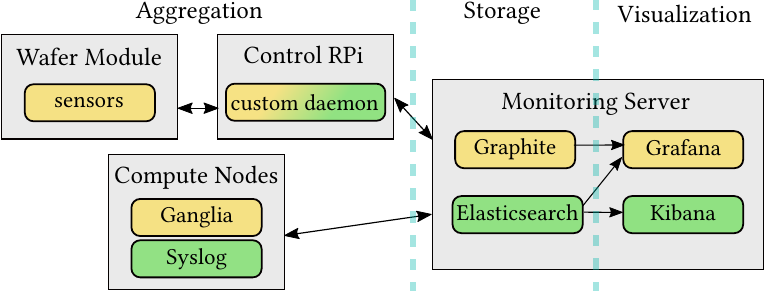}
	\caption{Flow of monitoring data from aggregation to storage and visualization.
             Grey boxes represent involved devices on which the corresponding software libraries run.
             Software responsible for time-series data is shown in yellow and for event data in green.
             Arrows illustrate connectivity between the different components.
             }
	\label{fig:wmod_monitoring}
\end{figure}

The general flow for monitoring data of a wafer module is shown in \cref{fig:wmod_monitoring}.
There are around 1200 time-series data sources within one wafer module.
Important sensors like wafer temperature are read out every few seconds.
Additionally, events for powering parts on/off or alerts are generated.
The data aggregation is performed on a Raspberry Pi via a software daemon handling several communications channels, e.g., I2C.
On the Raspberry Pi a first data analysis is done in order to have a quick response to a dangerous system state.
For example the temperatures are checked to be in an allowed range, above a given threshold the system is turned off.
Furthermore, all microcontrollers providing data to the Raspberry Pi perform local data evaluation tests and, therefore, detect false states faster than the Raspberry Pi.
Time-series data is stored on a central \codeInline{Carbon}~\cite{carbon} server outside the Raspberry Pi.
For the conventional compute nodes we use \codeInline{Ganglia}~\cite{massie2012monitoring} for data aggregation which also feeds into the \codeInline{Graphite} data base.
\codeInline{Graphite} uses a round-robin database for automatic data compression after certain intervals.
Event aggregation is done utilizing \codeInline{syslog}~\cite{syslog} which is parsed by \codeInline{Logstash}~\cite{logstash}.
Filtered events are stored in an \codeInline{Elasticsearch}~\cite{elasticsearch} database.
\codeInline{Grafana}~\cite{grafana} is used to visualized time-series data.
It allows the creation of dashboards which give insight to the state of the system on various levels of detail.
This facilitates getting a quick overview of relevant data from wafer modules and the state of the conventional compute cluster while simultaneously allowing to drill down for more details.
Additionally, events like powering on components can also be shown in Grafana to easily link events and changes in time-series data.
In general we use \codeInline{Kibana}~\cite{kibana} to visualize event data.

\subsection{Commissioning}

\subsubsection{Digital memory tests}\label{sec:blacklisting}
In large complex hardware systems, variations of individual components are inevitable.
As already mentioned in section \ref{sec:redman} the behavior of these components change due to varying hardware parameters such as supply voltage and might disturb the execution of experiments.
As a result, it is important to keep track of the state of the components and be aware of it during experiment execution.
This is achieved by digital memory tests that are executed after assembly as well as periodically.
Here, for a specific hardware configuration given by, e.g., supply voltage and clock frequency, each digital memory of every HICANN is read/write-tested with random values.
The results are compared and if a malfunctioning component is found it is flagged in the availability management database.
The database reflects the hierarchical structure of the hardware, so that always the largest functional unit that exclusively depends on the malfunctioning components is flagged, shown in \cref{fig:blacklisting}.
The information can then be extended individually for each experiment and stored by serializing the updated database to disk, which is typically an XML-based file format, cf.\ \cref{sec:redman}.
During experiment execution this data is then deserialized, which allows for skipping the unavailable components.
Besides the experiment execution, the digital memory tests are also used in continuous integration to monitor and store the state of the hardware.

\begin{figure}[tbp]
	\includegraphics[width=\columnwidth]{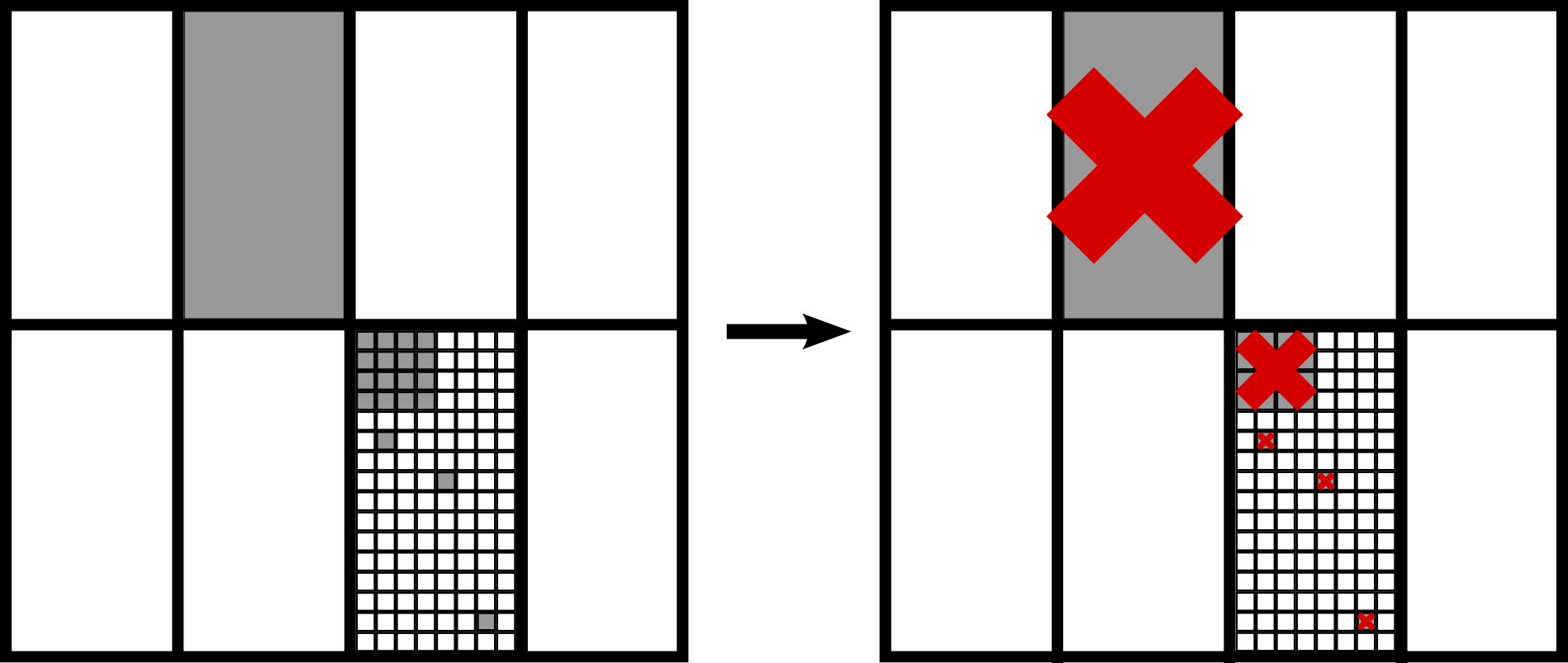}
	\caption{\label{fig:blacklisting}Digital test of malfunctioning components highlighted in grey.
	After the test (right side) these components are marked as not available using the hierarchy of the system.
	As a result, individual components up to large functional units, consisting of many components, are marked as not available.
	}
\end{figure}

\subsubsection{Calibration}
The one-time circuit characterization~\cite{githubcake} runs sequences of experiments that sweep the neuron parameters (stored as 10-bit values in analog floating gates on the HICANN), measures the changes' impacts, and employs different fits depending on the parameter effect's response.
A calibration database is then filled with the transformation data for its utilization on routine hardware usage.
The effect of applying such parameter mapping to the neuron configuration is exemplified in \cref{fig:calibration}.

\begin{figure}[tbp]
	\includegraphics[width=\columnwidth]{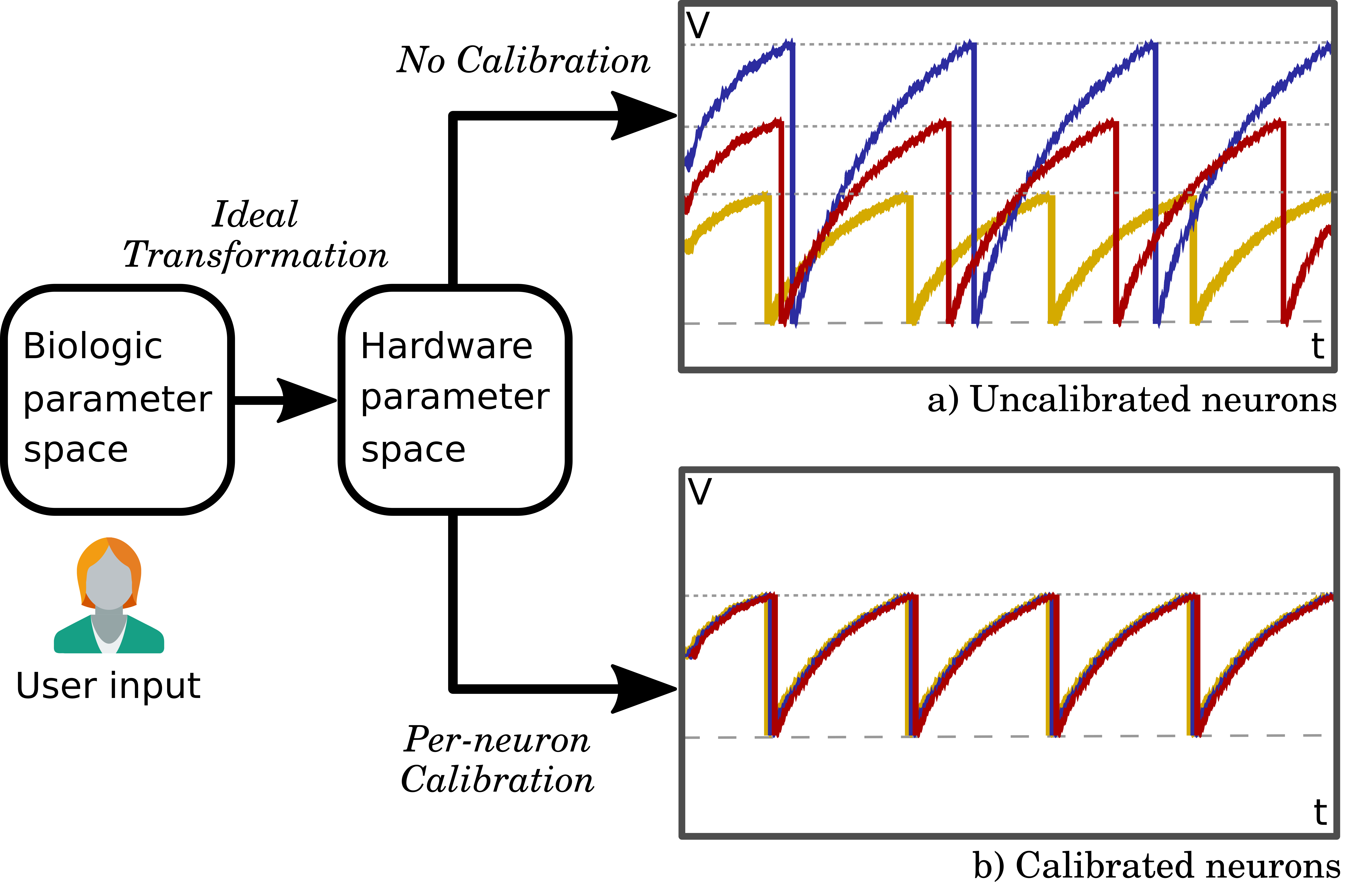}
	\caption{\label{fig:calibration}%
	Three neurons set for continuous spiking activity exhibit non-uniform threshold voltage under the same floating gate configuration.
	After applying the per-neuron calibration, parameters like $\protect V_{treshold}$ can be set accurately across different neuron circuits (Traces were hand-drawn for illustrative purposes, with attention only to $\protect V_{threshold}$).%
	}
\end{figure}

\section{Applications}\label{sec:app}

The previous sections motivated and detailed the status of the \stackName{}.
In the following, first a minimal experiment is demonstrated with the key concepts in action.
It is followed by examples for more complex ``full'' experiments.

\subsection{A Minimal Experiment}\label{sec:example_experiment}

\begin{listing}[tbp]
  \caption{Example for an experiment.}
\begin{minted}{python}
import numpy as np

# BrainScaleS OS imports
import pyhmf as pynn
import pyhalco_hicann_v2 as C
from pyhalco_common import Enum
import pyhalbe
import pysthal
from pysthal.command_line_util import init_logger
from pymarocco import PyMarocco
from pymarocco.runtime import Runtime
from pymarocco.results import Marocco

init_logger("WARN", [])

marocco = PyMarocco()
runtime = Runtime(C.Wafer(33))

pynn.setup(marocco=marocco, marocco_runtime=runtime)

experiment_duration = 1000  # ms

neuron_parameters = {
    'cm': 0.2,         # nF
    'v_reset': -20.,   # mV
    'v_rest': -20.,    # mV
    'v_thresh': 100,   # mV
    'e_rev_I': -20.,   # mV
    'e_rev_E': 0.,     # mV
    'tau_m': 10.,      # ms
    'tau_refrac': 0.1, # ms
    'tau_syn_E': 2.,   # ms
    'tau_syn_I': 5.,   # ms
}

pop = pynn.Population(2, pynn.IF_cond_exp, neuron_parameters)
stimulus = pynn.Population(1, pynn.SpikeSourceArray, {
    'spike_times': [0, 5, 10]})

# choose a specific HICANN for given pyNN population
marocco.manual_placement.on_hicann(pop, C.HICANNOnWafer(Enum(0)))

# record both, neuron membrane trace and spikes
pop.record()
pynn.PopulationView(pop).record_v()

# connect the stimulus
proj = pynn.Projection(stimulus, pop, pynn.AllToAllConnector(weights=0.001), target='excitatory')

# perform mapping but do not execute on hardware
marocco.backend = PyMarocco.None
pynn.run(experiment_duration)

# hardware-level settings
synapses_result = runtime.results().synapse_routing.synapses()
proj_items = synapses_result.find(proj)
for proj_item in proj_items:
    syn_c = proj_item.hardware_synapse()
    hicann = runtime.wafer()[syn_c.toHICANNOnWafer()]
    synapse = hicann.synapses[syn_c.toSynapseOnHICANN()]
    synapse.weight = pyhalbe.HICANN.SynapseWeight(3)

# run on hardware
marocco.skip_mapping = True
marocco.backend = PyMarocco.Hardware
pynn.run(experiment_duration)

np.savetxt("membrane.txt", pop.get_v())
np.savetxt("spikes.txt", pop.getSpikes())
\end{minted}
  \label{lst:experiment_example}
\end{listing}

\Cref{lst:experiment_example} shows an example experiment.
It demonstrates all software features discussed in the previous sections.
First, a couple of \codeInline{Python} modules are imported.
The \codeInline{marocco} object is instantiated that allows for custom configurations that are not part of the \codeInline{PyNN} API\@.
The \codeInline{runtime} object holds, amongst others, the \codeInline{sthal} representation of the wafer configuration that will be used for low-level re-configuration.
Next, parameters like the experiment duration and neuron parameters are set as variables.
A population of neurons as well as a stimulus are created with two and one neuron, respectively.
The population of neurons is placed on a specific HICANN\@.
If manual placement is not given, the mapping software will find a location depending on the chosen mapping algorithm.
No mapping hint is given for the stimulus.
It will be inserted as close as possible to the mapped neuron population while adhering to bandwidth limitations as good as possible.
The population is then asked to record both, its spikes and membrane potential.
Next, a projection is drawn between the stimulus and the neurons.
The projection is stored in a variable for later lookup.

Now that the network is completely setup, the mapping can be carried out, but it is not yet executed (\codeInline{backend=None}).
By this, the user can look up the hardware synapse between the stimulus and the neurons.
Doing so we set manually a digital weight of 3.
Then we skip the mapping, set the backend to hardware and execute.
After \codeInline{pynn.run} the resulting membrane trace and spikes can be read out.

\begin{listing}[htbp]
  \caption{Example for an experiment invocation.}
  \begin{minted}{shell}
  # allocate the full module
  srun -p experiment --wafer 33 experiment_example.py

  # allocate only HICANN 0 (with analog readout and trigger by default)
  srun -p experiment --wafer 33 --hicann 0 experiment_example.py
  \end{minted}
  \label{lst:experiment_example_run}
\end{listing}

The network execution is then invoked by calling \cref{lst:experiment_example} with a \codeInline{SLURM} command like \codeInline{srun}, see \cref{lst:experiment_example_run}.
The system on which the experiment is conducted is given as well as the partition which is used for accounting and priority.

\subsection{Examples for Full Experiments}

In the simple example explained above, the split between mapping and execution is only necessary if low-level access is wished.
However, an important application are chip-in-the-loop experiments where it is crucial that an iterative re-configuration is possible to, e.g., compensate for trial-to-trial variations.

\begin{figure}[tbp]
  \centering
  \includegraphics[width=\linewidth]{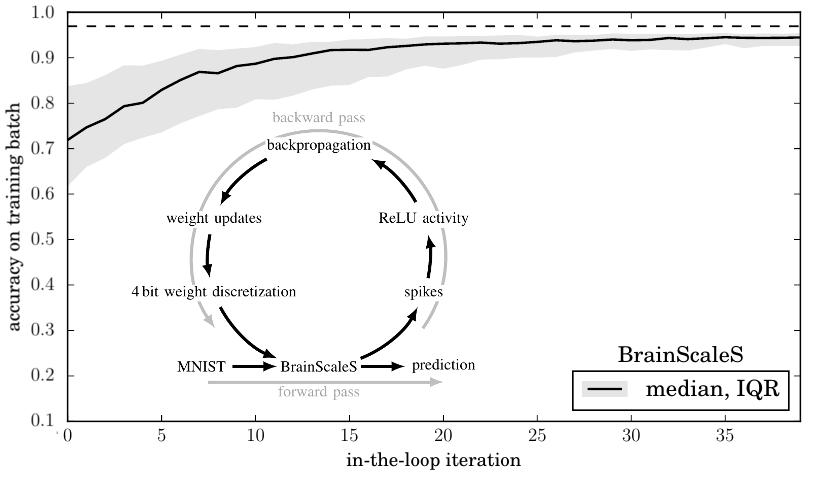}
  \caption{
        Each iteration of in-the-loop training consists of two passes.
        In the forward pass, the output firing rates of the LIF network are measured in hardware.
        In the backward pass, these rates are used to update the synaptic weights of the LIF network by computing the corresponding weight updates in the ReLU network and mapping them back to the hardware.
        Adapted from~\textcite{schmitt2016classification}.
  }
  \label{fig:itl}
\end{figure}

An experiment where this was used is detailed in~\cite{schmitt2016classification} for training of a deep network for digit classification.
\Cref{fig:itl} shows the concept.
After training an artificial neural network, the weights of a matching hardware network are set accordingly.
However, due to both, trial-to-trial variations and differing responses of the artificial w.r.t.\ the hardware neurons, the classification performance is diminished.
By continuing the training in the loop, the performance can be restored.
For this, the response of all neurons in the hardware network is fed back into the training loop of the artificial network.

Accelerated physical emulation of Bayesian inference in spiking neural networks was demonstrated in~\textcite{kungl2019accelerated} where the full \stackName{} was used as well.
A network of spiking neurons was set up to sample from a Boltzmann distribution.
The network was also trained iteratively, however, not in companion with an artificial network.
Classification and pattern completion were demonstrated on two datasets.

Another example is~\textcite{goeltz2019fast} that demonstrates classification based on spike timing only.
Again, the hardware in-the-loop approach was used to train a network classifying images on \BSS{1}.

\section{Future Developments}\label{sec:future}

\subsection{Separation of Experiment Configuration and Execution }

The experiment demonstrated in \cref{sec:example_experiment} executed the mapping and the neuromorphic emulation in one process.
Most importantly, the requested hardware resources had to be specified prior to the mapping.
This is an unfortunate order as it does not allow for, e.g., choosing the system dynamically or specifying only a subset of the wafer for running experiments in parallel.
However, the necessary ingredients to overcome this problem are in place and work is currently carried out to implement a solution based on the serialization capabilities of the data structures, see \cref{sec:serialization}.

\subsection{Next-generation \codeInline{Python} Binding Generation}

The design and development of the \stackName started in 2009.
In the meantime, several external dependencies have been deprecated.
In particular, our auto-generated \codeInline{Python} wrapping depends on \codeInline{gccxml} where development stopped in 2015.
It depends on \codeInline{gcc} $\le 4.9.3$ blocking the usage of the latest \codeInline{C++} features from the 14, 17 and 20 standards in header files.
We evaluated several approaches, including the usage of \codeInline{LLVM}'s \codeInline{castxml}, but resorted to developing a new wrapper code generator ---\codeInline{genpybind} \cite{githubgenpybind}--- which is based on \codeInline{LLVM} libraries.
The transition from the \codeInline{py++}-based to \codeInline{genpybind} is now in progress.
In addition, \codeInline{genpybind} also attacks binding generation from a different angle with a more fine-grained and explicit approach.

\subsection{Towards \protect\BrainScaleS{2}}

Software development for \BSS{2} started in 2016 and builds upon the results ---the \stackName--- presented in this work.
We try to re-use and adapt as much as possible from the existing code base.
Especially the coordinate system, cf.\ \cref{par_coord}, has proven beneficial.
However, early in the design phase we decided to improve the hardware abstraction layers by introducing structured types encapsulating all on-chip and on-FPGA configurable hardware entities.
These types also provide explicit implementations for encoding to and decoding from hardware configuration bitstreams.
In conjunction with a timer-based execution flow on the FPGA, this allows for experiments being described as an timed event sequence.
The \codeInline{C++} API makes heavy use of \codeInline{std::future}-like interfaces to expose an asynchronous interface to the experiment control flow.
Additionally, a fast experiment scheduler has been developed for \BSS{2}.
It allows for approximately ten experiments per second where ---due to the neuromorphic speed-up factor--- each experiment represents up to \SI{100}{\second} of emulated time in the biological model.
Based on this, restructuring work on \BSS{1} started and a corresponding implementation was developed.
Similarly, the \codeInline{genpybind} tool was created during \BSS{2} low-level software development.
See~\textcite{mueller2020bss2ll} for a detailed description.

\section{Discussion}\label{sec:discussion}

This work describes the latest version of \stackName, the software stack operating the \BSS{1} platform.
It allows to accomplish the main goal of the wafer-scale mixed-signal neuromorphic system \BrainScaleS{1}:
designing and running wafer-scale experiments.
The software stack aims for non-expert usage, e.g., by neuroscientists, while maintaining access to all other abstraction levels for expert users.
We give a detailed overview of the individual software components and describe different aspects.
From the hardware configuration, over the interaction with the system, e.g., setup, runtime control and result read out.
We describe the transformation of user-defined experiments into a valid hardware configuration, as well as the necessary resource management and monitoring.

BrainScaleS adopted development methodologies and tools originating in software engineering to improve platform robustness and experiment reproducibility.
\BSS{1} is operated as a platform which is available for the research community.

Several experiments~\cite{schmitt2017hwitl,kungl2019accelerated,goeltz2019fast} demonstrate that \stackName{} is a viable basis for using \BSS{1}.
In addition to the publications, several thesis in our group made use of it for conducting neuromorphic experiments and commissioning work.

\section{Contributions}

E.~Müller is the lead developer and architect of the BrainScaleS software stack.
S.~Schmitt contributed to the calibration, the stateful configuration layer and the general usage flow.
C.~Mauch contributed to the system configuration layers as well as system operation.
S.~Billaudelle is a main contributor to the \BSS{1} \codeInline{PyNN} API implementation.
A.~Grübl contributed software for low-level configuration as well as the system simulation backend.
M.~Güttler is the main developer of the system-level operation software, e.g., system monitoring and controlling, and contributed to low-level firmware.
D.~Husmann is the main developer of the system-level test software suite and contributed to system-level operation software.
J.~Ilmberger contributed to the communication layer and the analog readout framework.
S.~Jeltsch is the main developer of the map \& route layer.
J.~Kaiser contributed to speed-up the synapse configuration.
J.~Klähn contributed to the map \& route layer.
M.~Kleider contributed to the calibration of the system.
C.~Koke is the main developer of the stateful configuration layer and the calibration framework.
J.~Montes contributed to calibration scalability.
P.~Müller evaluated the performance of the \BSS{1} neuromorphic circuit implementation.
J.~Partzsch contributed software for the low-level system configuration.
F.~Passenberg optimized the map \& route algorithms to enable successful topology mapping to wafer modules with non-ideal blacklisting state.
H.~Schmidt contributed to digital blacklisting.
B.~Vogginger is a main developer to the simulation backend and contributed to the map \& route layer.
J.~Weidner contributed to the web-based configuration visualization and acquired configuration results for the Jülich cortical column network. %
C.~Mayr contributed to the system design (hardware and software) of the off-wafer communication stack.
J.~Schemmel is the lead designer and architect of the \BrainScaleS{1} neuromorphic system.
All authors discussed and contributed to the manuscript.

\section*{Acknowledgments}

The authors wish to thank all present and former members of the Electronic Vision(s) research group contributing to the BSS-1 hardware system, development and operation methodologies, as well as software development.
The authors express their special gratitude towards:
\begin{inparaenum}
\item Daniel Barley for his contribution to the parallel ADC-readout software;
\item Richard Boell for his contribution to the web-based visualization tool;
\item Patrick Häussermann for his contribution to experiment isolation in the scheduler;
\item Kai Husmann for his contribution to the low-level system control environment;
\item Lukas Pilz for his contribution to evaluate support for iterative configuration;
\item Vitali Karasenko for his contribution to the graphics;
\item Alexander Kononov for his effort when commissioning the HICANN chip;
\item Daniel Kutny for his contribution to the monitoring solution;
\item Dominik Schmidt for his contribution to the calibration framework;
\item Moritz Schilling for his contribution to the initial implementation of the secure transport layer protocol;
\item Andreas Baumbach,
\item Oliver Breitwieser and
\item Yannik Stradmann for their work on continuous integration and structured deployment of the software environment.
\end{inparaenum}
We especially express our gratefulness to the late Karlheinz Meier who initiated and led the project for most if its time.

This work has received funding from the EU ([FP7/2007-2013], [H2020/2014-2020]) under grant agreements 604102 (HBP), 269921 (BrainScaleS), 243914 (Brain-i-Nets), 720270 (HBP SGA1), 785907 (HBP SGA2) and 945539 (HBP SGA3), the \mbox{Deutsche} \mbox{Forschungsgemeinschaft} (DFG, German Research Foundation) under Germany’s Excellence Strategy EXC 2181/1-390900948 (the Heidelberg STRUCTURES Excellence Cluster), the Helmholtz Association Initiative and Networking Fund [Advanced Computing Architectures (ACA)] under Project SO-092, as well as from the Manfred Stärk Foundation.

\ifoptionfinal{\IEEEtriggeratref{60}}{}
\printbibliography[notkeyword=own_software]
\printbibliography[title={Own Software},keyword=own_software]

\end{document}